\documentclass[letterpaper, 10 pt, conference]{ieeeconf}  

\IEEEoverridecommandlockouts                              

\usepackage{amsmath,amssymb}
\usepackage{booktabs}
\usepackage{multirow}
\usepackage{graphicx}
\usepackage{subfig}
\usepackage[hidelinks]{hyperref}
\usepackage{cleveref}
\usepackage{algorithm}
\usepackage{algpseudocode}
\usepackage{xcolor}
\usepackage{stfloats}
\title{\LARGE \bf
VOKE: Visual Object Kinematics Estimator for Dexterous In-Hand Manipulation
}

\author{Shengjia Zhu$^{1,2,*}$, Wenqiang Xu$^{3}$, and Lin Shao$^{2}$%
\thanks{This work has been submitted to the IEEE for possible publication. Copyright may be transferred without notice, after which this version may no longer be accessible.}%
\thanks{$^{1}$Shanghai Jiao Tong University.}%
\thanks{$^{2}$National University of Singapore.}%
\thanks{$^{3}$Fudan University.}%
\thanks{$^{*}$This work was conducted while Shengjia Zhu was a student at Shanghai Jiao Tong University and a research intern at the National University of Singapore. Shengjia Zhu is now at The University of Texas at Austin.}%
}

\begin{document}

\maketitle
\thispagestyle{empty}
\pagestyle{empty}

\begin{abstract}
Vision-based dexterous in-hand manipulation requires knowing not only where an object is, but also how it moves. However, most robotic manipulation systems rely on off-the-shelf visual perception models whose outputs are limited to static object masks, geometry, or 6D poses, leaving downstream controllers to infer object dynamics implicitly from noisy visual observations. We present VOKE, a kinematics-aware perception framework that augments existing object and hand-object pose estimators to recover a complete object kinematic state comprising pose, linear and angular velocity, and linear and angular acceleration. VOKE contains two core perception modules and a downstream application study. First, a \textbf{Pluggable Kinematics Module (PKM)} attaches to the output of a standard pose estimator and predicts the object's linear and angular velocity and acceleration from temporal visual features. Second, a \textbf{Kinematics-guided Pose Refinement Module (KPRM)} converts this motion vector into a second-order, motion-prior reference pose and refines the per-frame pose estimates. We evaluate VOKE on DexYCB and OakInk, where it consistently improves object pose estimation across three different backbones and substantially reduces velocity and acceleration estimation error compared with pose differencing. As a downstream application, we use \textbf{Kinematics-Augmented Reinforcement Learning (KARL)} to compare a kinematics-augmented policy with a pose-history baseline. After training the policies in simulation, we transfer the kinematics-augmented policy to the physical robot without policy fine-tuning, demonstrating the complete perception-to-control pathway in the target domain. More results and visualizations are available at \nolinkurl{https://voke-project.pages.dev/}.
\end{abstract}

\begin{figure}[!t]
  \centering
  \includegraphics[width=\columnwidth]{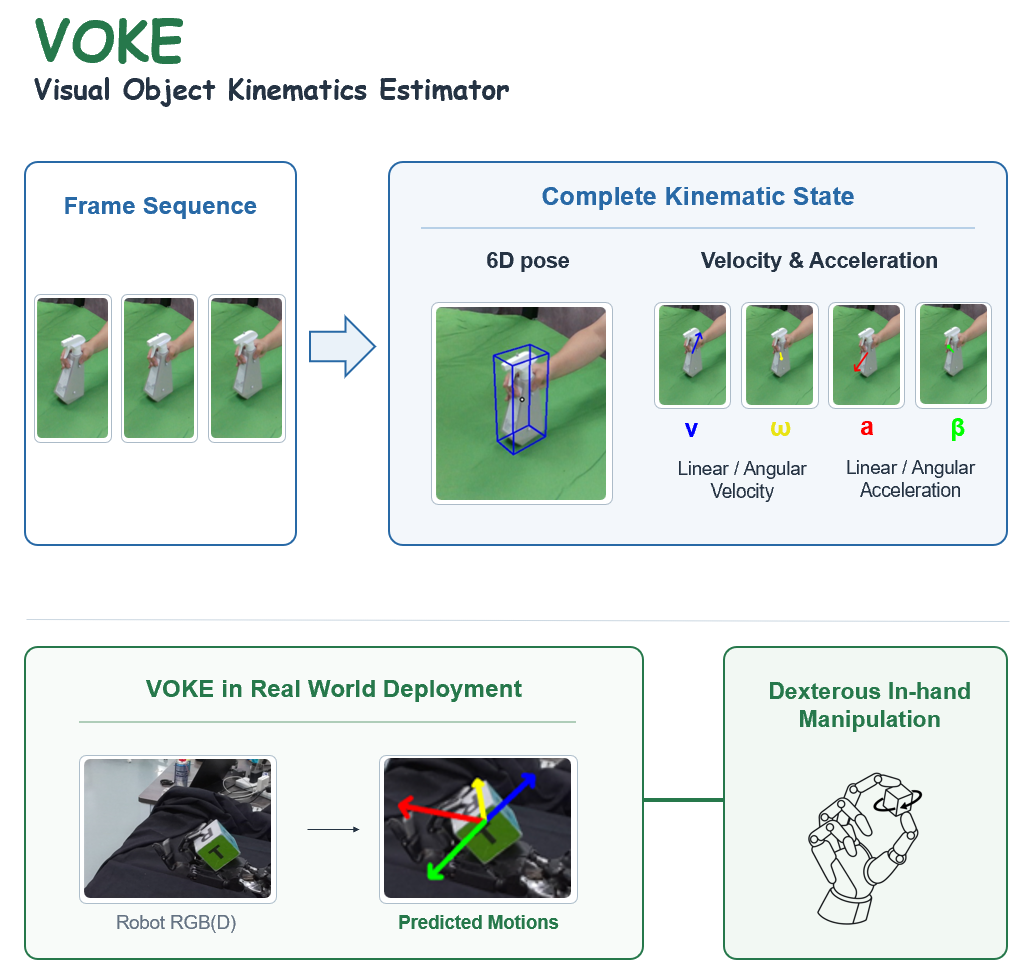}
  \caption{Motivation and overview of VOKE. Top: VOKE augments frame-wise object pose estimation with explicit linear and angular velocity and acceleration. Bottom: During real-world deployment, VOKE converts live robot RGB(D) observations into kinematics estimates for dexterous in-hand manipulation.}
  \label{fig:intro_overview}
\end{figure}

\section{Introduction}
	
Estimating an object's complete kinematic state---its 6D pose together with linear and angular velocity and acceleration---from visual data is a crucial step toward enabling more physically informed robotic manipulation. This is especially true for contact-rich tasks such as dexterous in-hand manipulation, where the object is frequently occluded and its motion changes rapidly. However, this problem is challenging for two main reasons. First, modern pose-estimation and tracking systems typically expose object pose rather than a complete kinematic state~\cite{foundationpose,megapose}, leaving downstream modules to infer velocity and acceleration from observation histories. Second, directly computing higher-order derivatives from noisy, image-based pose estimates is difficult because differentiation amplifies measurement noise, particularly for acceleration~\cite{savitzky1964smoothing}. We address these challenges with a kinematics-aware perception framework that makes the complete object kinematic state an explicit, first-class output of the visual pipeline.

To improve robustness against occlusion and temporal instability, previous studies have leveraged temporal relationships in videos to refine pose estimates or construct temporally consistent supervision~\cite{choi2021beyond,liu2021interactions}. These methods use motion-related cues without explicitly outputting the object's complete kinematic state. This valuable physical information is therefore unavailable to downstream control tasks. Our work fills this gap by designing a lightweight, pluggable module that predicts velocity and acceleration directly from temporal visual features and uses these estimates for both pose refinement and policy learning.

Our framework, VOKE (Visual Object Kinematics Estimator), operates on a sequence of RGB(D) hand-object observations, as illustrated in Fig.~\ref{fig:intro_overview}. Its perception core contains two modules. First, a \textbf{Pluggable Kinematics Module (PKM)} attaches to an object or hand-object pose estimator: it shares the image encoder of the base model, aggregates the features of the last $k$ frames, and regresses the motion vector $V_t=(v_t,\omega_t,a_t,\beta_t)$, containing linear and angular velocity and acceleration. Second, a \textbf{Kinematics-guided Pose Refinement Module (KPRM)} integrates $V_t$ into a second-order, motion-prior reference pose and refines the per-frame pose with a small MLP-based pose refiner. The refined pose and motion vector jointly form the complete kinematic state $K_t=(P_{t,\mathrm{fin}},V_t)$. We further use \textbf{Kinematics-Augmented Reinforcement Learning (KARL)} as a downstream application protocol in which a manipulation policy receives $K_t$ as part of its observation.

We evaluate VOKE primarily as a perception method. On the DexYCB~\cite{dexycb} and OakInk~\cite{oakink} benchmarks, we integrate VOKE into three representative pose estimators (gSDF~\cite{gsdf}, HFL-Net~\cite{hflnet}, and HOISDF~\cite{hoisdf}) and measure object pose and reconstruction quality; we additionally report the accuracy of the predicted velocity and acceleration as a reference baseline for future work on visual object kinematics estimation. As a downstream application, we train a dexterous in-hand cube reorientation policy for a Shadow Dexterous Hand in Isaac Sim. The kinematics-augmented policy converges faster during training. With VOKE-predicted object states, it achieves a higher success rate (78.90\% vs.\ 68.20\%) and a lower final rotation error (8.90 vs.\ 12.75 degrees) than the pose-history baseline. We then deploy the frozen simulation-trained policy on the physical robot, where real-time VOKE predictions replace the simulator object state to form the policy observation and close the perception-to-control loop.

Our contributions are threefold:
\begin{itemize}
    \item A pluggable kinematics module (PKM) that endows existing pose estimators with explicit velocity and acceleration prediction and supports temporally informed pose refinement.
    \item A kinematics-guided pose refinement module (KPRM) that introduces a second-order motion prior between consecutive frames and improves frame-wise pose estimates.
    \item A systematic empirical study on DexYCB, OakInk, and a simulated dexterous in-hand manipulation dataset, including quantitative reference baselines for image-based object velocity/acceleration estimation in hand-object interaction, a simulated study of kinematics-augmented RL (KARL), and real-world deployment in which online VOKE predictions provide object-state inputs to a simulation-trained robot policy.
\end{itemize}


\section{Related Work}
\subsection{Object and Hand-Object Pose Estimation}

Estimating the 6D pose of objects from RGB or RGB-D images is a long-standing problem. General-purpose methods estimate and track novel-object 6D pose by combining learned representations with rendering, correspondence matching, and temporal tracking~\cite{foundationpose,megapose}. In hand-object interaction, HOPE-Net~\cite{doosti2020hope}, gSDF~\cite{gsdf}, HFL-Net~\cite{hflnet}, and HOISDF~\cite{hoisdf} estimate hand and object poses or reconstruct their geometry from monocular RGB images under mutual occlusion. The configurations evaluated here produce pose or geometry estimates without explicitly exposing object velocity or acceleration; VOKE complements them with an explicit kinematic output.

\subsection{Temporal and Motion-Aware Pose Estimation}

Temporal information has been used to improve pose consistency in several related settings. Choi \emph{et al.} refine per-frame human pose and shape estimates using motion features learned from video~\cite{choi2021beyond}, while Liu \emph{et al.} use spatial and temporal consistency to obtain pseudo-labels for semi-supervised hand-object pose estimation~\cite{liu2021interactions}. Among rigid-object trackers, 6-PACK propagates category-level poses using anchor-based keypoints~\cite{wang2020sixpack}, whereas BundleSDF jointly tracks and reconstructs unknown objects~\cite{wen2023bundlesdf}. More directly related to kinematics estimation, ROFT combines RGB-D pose observations with optical flow to track object pose and velocity~\cite{piga2022roft}, while TEG-Track extracts tactile kinematic cues, including object velocity, to improve 6D pose tracking of an in-hand object~\cite{liu2024tegtrack}. Task-specific systems also estimate per-object velocity in monocular driving videos~\cite{brazil2020kinematic}. These methods estimate motion in different domains or sensing settings, but none outputs all four motion quantities considered here—linear and angular velocity and acceleration—from temporal visual features.

Object kinematics also serve as intermediate quantities in physics-based analyses of manipulation. Pham \emph{et al.} use markerless visual tracking and hand-object dynamics to estimate contact forces~\cite{pham2018contact}, whereas Li \emph{et al.} recover person-object motion and interaction forces through trajectory optimization with contact and dynamics constraints~\cite{li2019motionforces}. These studies motivate visual motion estimates but address different objectives and inference pipelines.

\subsection{Perception for Dexterous Manipulation}

Dexterous in-hand manipulation has benefited from large-scale reinforcement learning and domain randomization for sim-to-real transfer~\cite{andrychowicz2020inhand}. A complementary direction combines tactile and proprioceptive state estimation with reinforcement learning~\cite{pitz2023dextrous}. The policy observations used by these systems vary across tasks and may include simulator states, estimated object poses, tactile measurements, or observation histories. Model-based methods instead learn action-conditioned dynamics and use model-predictive control for dexterous in-hand manipulation~\cite{nagabandi2020deep}. Our application study asks whether an explicit, visually estimated kinematic state is useful as a policy observation.

\section{Method}

\begin{figure*}[!t]
  \centering
  \includegraphics[width=0.9\linewidth]{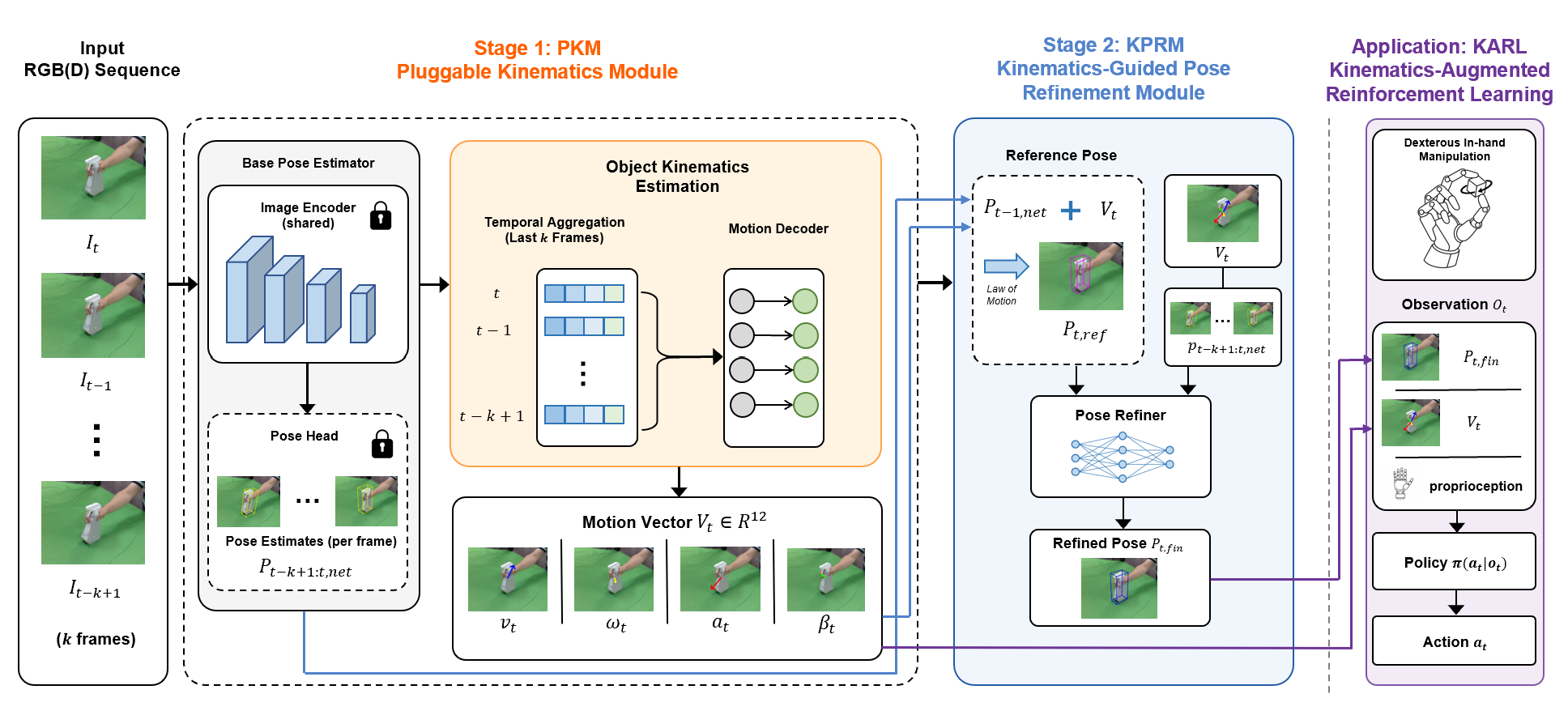}
  \caption{Overview of VOKE. Given a sequence of RGB(D) frames, a base pose estimator shares its image encoder with the Pluggable Kinematics Module (PKM) and produces per-frame pose estimates. PKM aggregates the last $k$ frame features and decodes the motion vector $V_t=(v_t,\omega_t,a_t,\beta_t)$. The Kinematics-guided Pose Refinement Module (KPRM) constructs a motion-prior reference pose from $P_{t-1,\mathrm{net}}$ and $V_t$, then refines the per-frame pose using $V_t$ and the pose history $P_{t-k+1:t,\mathrm{net}}$. The refined pose and motion vector form the complete kinematic state $K_t=(P_{t,\mathrm{fin}},V_t)$ consumed by the Kinematics-Augmented Reinforcement Learning (KARL) policy together with robot proprioception.}
  \label{fig:overview}
\end{figure*}

This section presents the VOKE perception framework and its downstream application to dexterous in-hand manipulation (Fig.~\ref{fig:overview}). Given a sequence of RGB(D) images, a base pose estimator first produces per-frame object poses. Building on these estimates, the \textbf{Pluggable Kinematics Module (PKM)} predicts the object's linear and angular velocity and acceleration from temporal visual features (Sec.~\ref{sec:pkm}). The \textbf{Kinematics-guided Pose Refinement Module (KPRM)} then integrates this motion vector into a second-order, motion-prior reference pose and refines the per-frame pose (Sec.~\ref{sec:kpr}). Together, the refined pose and motion vector constitute the complete kinematic state used by \textbf{Kinematics-Augmented Reinforcement Learning (KARL)} (Sec.~\ref{sec:karl}).

\subsection{PKM: Pluggable Kinematics Module}
\label{sec:pkm}

In hand-object or object pose estimation, a standard model independently predicts the object's translation $p \in \mathbb{R}^3$ and rotation $R \in SO(3)$ in the camera frame for each frame. To exploit motion consistency across frames, PKM processes the last $k$ frames $I_{t-k+1:t}$ to estimate the motion vector $V_t \in \mathbb{R}^{12}$, consisting of linear velocity $v_t$, angular velocity $\omega_t$, linear acceleration $a_t$, and angular acceleration $\beta_t$, each in $\mathbb{R}^3$.

PKM is designed to be model-agnostic. It shares the image encoder of the base pose estimator and concatenates the features of the last $k$ frames to aggregate temporal information. The concatenated features $f_{t-k+1:t}$ are passed through a lightweight MLP-based motion decoder to produce $V_t=(v_t,\omega_t,a_t,\beta_t)$. The head is trained with a Huber loss~\cite{huber1964robust} against the motion-quantity reference values described in Sec.~\ref{sec:datasets}. Because PKM reads temporal visual features rather than differentiating the estimated pose sequence, it can exploit appearance-level motion cues while leaving the base estimator unchanged.



\subsection{KPRM: Kinematics-Guided Pose Refinement Module}
\label{sec:kpr}

We use the estimated motion vector to construct a local motion prior for pose refinement. We compute a reference pose for frame $t$ from the previous network estimate $P_{t-1,\mathrm{net}}=(p_{t-1,\mathrm{net}},R_{t-1,\mathrm{net}})$ and the current motion vector $V_t=(v_t,\omega_t,a_t,\beta_t)$:
\begin{equation}
\begin{aligned}
p_{t,\mathrm{ref}} &= p_{t-1,\mathrm{net}} + v_{t}\Delta t - \frac{1}{2}a_{t}{\Delta t}^2, \\
R_{t,\mathrm{ref}} &= R_{t-1,\mathrm{net}}\,\mathrm{Exp}\!\left[\left(\omega_{t}-\frac{1}{2}\beta_t\Delta t\right)\Delta t\right],
\end{aligned}
\label{eq:kinematics}
\end{equation}
where $\Delta t$ is the frame interval and $\mathrm{Exp}:\mathbb{R}^3\rightarrow SO(3)$ maps a rotation vector to a rotation matrix. Since $v_t$ and $\omega_t$ are endpoint velocities, subtracting half of the acceleration increment approximates the interval-average velocity. Equation~\eqref{eq:kinematics} is used as a short-horizon second-order prior, not as an exact dynamics constraint.

We use a lightweight MLP-based pose refiner that takes the reference pose $P_{t,\mathrm{ref}}$, the estimated motion vector $V_t$, and the original pose estimates of the last $k$ frames $P_{t-k+1:t,\mathrm{net}}$ as input and outputs the refined pose $P_{t,\mathrm{fin}}$:
\begin{equation}
P_{t,\mathrm{fin}} = f(P_{t,\mathrm{ref}}, V_t, P_{t-k+1:t,\mathrm{net}}),
\label{eq:pose_refiner}
\end{equation}
where $f$ denotes the pose refiner. Because the reference pose carries temporal information through an explicit physics-inspired motion prior, the refined pose is more temporally consistent and accurate. We supervise the refined pose by penalizing the $L_2$ error of the object's eight bounding box corners. Together, PKM and KPRM produce the complete kinematic state $K_t=(P_{t,\mathrm{fin}},V_t)$ consumed by downstream policies, regardless of the base pose estimator.

\subsection{KARL: Kinematics-Augmented Reinforcement Learning}
\label{sec:karl}

We formulate the dexterous in-hand manipulation task as a Markov Decision Process $\mathcal{M}=(S,O,A,\pi,\tau,R)$, where $S$, $O$, and $A$ are the state, observation, and action spaces, $\pi$ is the policy, $\tau$ is the transition distribution, and $R$ is the reward. With VOKE, the policy observes the complete kinematic state $K_t=(P_{t,\mathrm{fin}},V_t)$ in addition to standard proprioceptive and goal-related inputs. This object-centric dynamic observation provides a more complete approximation of the Markov state than a static pose alone and is expected to improve policy robustness. An application study of KARL is provided in Sec.~\ref{sec:application}.



\section{Experiments}
Our evaluation is centered on perception: we primarily verify (i) whether VOKE improves object pose estimation and (ii) whether it estimates object velocity and acceleration accurately, providing reference baselines for future work. A dexterous in-hand manipulation task is then presented as an application of the full design. The experiments are organized to answer the following questions:
\begin{itemize}
    \item \textbf{Q1.} Does integrating PKM and KPRM improve the accuracy and temporal consistency of object pose estimation across different backbones?
    \item \textbf{Q2.} Can VOKE estimate object velocity and acceleration accurately from visual observations under complex hand-object interaction?
    \item \textbf{Q3.} Can VOKE serve as a practical perception-to-control interface for dexterous in-hand manipulation?
\end{itemize}




\subsection{Datasets and Evaluation Setup}
\label{sec:datasets}

\subsubsection{Datasets and Preprocessing}

\textbf{DexYCB}~\cite{dexycb} is a widely used dataset for hand-object pose estimation. It contains 582K RGB-D frames across 1,000 sequences of 10 subjects grasping 20 objects, captured from 8 views. We use right-hand samples and the official $s_0$ split. We remove frames without hand-object interaction or with most of the hand or object outside the image.

\textbf{OakInk}~\cite{oakink} is a dataset for visual and cognitive understanding of hand-object interactions. It contains 230K RGB-D frames of 12 subjects interacting with 100 object instances from 32 categories. We apply the same filtering as for DexYCB.

\textbf{Dexterous in-hand manipulation simulation dataset.} To evaluate VOKE under dexterous in-hand manipulation scenarios, we construct a synthetic dataset in Isaac Sim from the \textit{Shadow Hand Repose Cube} task in Isaac Lab~\cite{mittal2025isaaclab}. We record 40 trajectories of a trained cube-reorientation policy, each containing 300 frames at 30 FPS and rendered from 50 camera views, for approximately 600K view--frame observations. We split the dataset by complete trajectories, using 32 trajectories for training and 8 for validation.

\textbf{Motion-quantity reference values.} DexYCB and OakInk do not provide measured object velocity or acceleration, so we derive reference values from their ground-truth object-to-camera pose trajectories and timestamps. We fit a smoothed cubic spline to each translation trajectory and evaluate its first and second analytic derivatives at time $t$ to obtain $v_t$ and $a_t$. We compute $\omega_t$ by applying a centered difference to adjacent relative rotations mapped to the local $SO(3)$ tangent space, and obtain $\beta_t$ by differentiating a smoothed cubic spline fitted to the resulting angular-velocity sequence.

For the Isaac Sim dataset, object pose, linear velocity, and angular velocity are provided directly by the simulator. We obtain linear and angular acceleration by fitting smoothed cubic splines to the corresponding simulator-provided velocity sequences and taking their analytic first derivatives.

To verify that the derived kinematics are physically meaningful, we replay the object motion in MuJoCo~\cite{todorov2012mujoco} with external forces and torques computed from the kinematics; the replayed trajectories align well with the annotated trajectories.

\subsubsection{Evaluation Metrics}

\textbf{Pose estimation.} Following the conventions of the three backbones, we report the object center error ($E_o$, cm), mean corner error ($E_c$, cm), ADD-0.1D(s)(\%)~\cite{hflnet}, ADD-S (cm)~\cite{hoisdf}, object-mesh Chamfer Distance ($CD_o$, cm$^2$), and object F-score at 5 and 10 mm ($FS_o@5$ and $FS_o@10$)~\cite{gsdf}.

\textbf{Temporal stability of pose predictions.} In addition to frame-wise pose accuracy, we report one-step translational and rotational Relative Pose Error (RPE) following Sturm \emph{et al.}~\cite{sturm2012benchmark}. Translation and rotation are reported in centimeters and degrees, respectively; lower is better.

\textbf{Velocity and acceleration estimation.} For each $x \in \{v,\omega,a,\beta\}$, mean Euclidean error (MEE) averages $\|\hat{x}_t-x_t\|_2$ over frames. Normalized root mean squared error (NRMSE) divides the root mean squared vector error by the root mean square magnitude of the reference vectors. MEE is reported in $m/s$ for linear velocity, $m/s^2$ for linear acceleration, $rad/s$ for angular velocity, and $rad/s^2$ for angular acceleration; NRMSE is dimensionless.

\subsection{Main Results: Object Pose Estimation}
\label{sec:main_pose}

We integrate VOKE into three representative hand-object pose estimation and reconstruction baselines: gSDF~\cite{gsdf}, HFL-Net~\cite{hflnet}, and HOISDF~\cite{hoisdf}. For each baseline, we attach PKM and KPRM to the original pipeline and compare the refined object pose with the original estimate.

\textbf{gSDF}~\cite{gsdf} is a geometry-driven hand-object reconstruction framework that estimates hand and object geometry from a monocular RGB image using signed distance functions. The released configuration used in our experiments provides an object-center estimate but no explicit rotation output, so we add a rotation branch to obtain the complete object pose required by VOKE. \textbf{HFL-Net}~\cite{hflnet} jointly predicts the hand pose and the object's 6D pose from a single RGB image using harmonious feature learning. \textbf{HOISDF}~\cite{hoisdf} introduces global signed distance fields as intermediate 3D representations to constrain hand and object pose regression.

The frame-wise pose results and the temporal stability of the pose predictions are reported in Table~\ref{tab:main_pose}. ``--'' marks entries unavailable from those outputs. Across all three backbones and both datasets, integrating PKM and KPRM improves every available pose and reconstruction metric, showing that the gains are not tied to a particular representation or evaluation measure. VOKE also improves temporal stability for every evaluated backbone--dataset pair, reducing translational RPE by 41.0--64.4\% and rotational RPE by 7.9--33.3\%. These consistent spatial and temporal gains answer \textbf{Q1} affirmatively.

\begin{table*}[!t]
\caption{Main object-pose estimation results and temporal stability of the pose predictions on DexYCB and OakInk. Values absent from the original publications are recomputed from our inference outputs where possible; ``--'' denotes metrics unavailable from those outputs. Here, $E_o$, $E_c$, and ADD-S are in cm; ADD-0.1D(s) is in percent; $CD_o$ is in cm$^2$; and translational and rotational RPE are in cm and degrees, respectively.}
\label{tab:main_pose}
  \centering
  \resizebox{\textwidth}{!}{
  \begin{tabular}{llcccccccccc}
    \toprule
    Backbone & Method & $E_o\downarrow$ & $E_c\downarrow$ & ADD-S$\downarrow$ & ADD-0.1D(s)$\uparrow$ & \multicolumn{2}{c}{$CD_o\downarrow$} & $FS_o@5\uparrow$ & $FS_o@10\uparrow$ & \multicolumn{2}{c}{RPE$\downarrow$} \\
    \cmidrule{7-8}\cmidrule{11-12}
    & & & & & & mean & median & & & Trans. & Rot. \\
    \midrule
    \multirow{4}{*}{gSDF}
      & DexYCB \;\; orig.       & 1.95          & 2.89 & 1.40 & 52.5           & 3.76          & 1.58          & 0.439          & 0.707          & 2.50 & 4.90 \\
      & DexYCB \;\; +VOKE       & \textbf{1.80} & \textbf{2.75} & \textbf{1.35} & \textbf{53.3}           & \textbf{3.50} & \textbf{1.43} & \textbf{0.456} & \textbf{0.724} & \textbf{0.89} & \textbf{4.47} \\
      & OakInk \;\; orig.       & 2.24          & 3.15 & 1.55 & 48.9           & 6.51          & 1.52          & 0.414          & 0.677          & 2.65 & 5.04 \\
      & OakInk \;\; +VOKE       & \textbf{2.12} & \textbf{2.99} & \textbf{1.48} & \textbf{50.2}           & \textbf{6.07} & \textbf{1.46} & \textbf{0.428} & \textbf{0.686} & \textbf{1.06} & \textbf{4.64} \\
    \midrule
    \multirow{4}{*}{HFL-Net}
      & DexYCB \;\; orig.       & 3.98 & 5.57 & 3.19 & 30.2          & -- & -- & -- & -- & 3.67 & 4.45 \\
      & DexYCB \;\; +VOKE       & \textbf{3.78} & \textbf{5.40} & \textbf{3.01} & \textbf{32.0} & -- & -- & -- & -- & \textbf{2.02} & \textbf{3.93} \\
      & OakInk \;\; orig.       & 4.46 & 6.12 & 3.51 & 25.6 & -- & -- & -- & -- & 3.48 & 4.65 \\
      & OakInk \;\; +VOKE       & \textbf{4.29} & \textbf{5.95} & \textbf{3.40} & \textbf{27.8} & -- & -- & -- & -- & \textbf{1.80} & \textbf{3.10} \\
    \midrule
    \multirow{4}{*}{HOISDF}
      & DexYCB \;\; orig.       & 1.84          & 2.74          & 1.33          & 58.3 & -- & -- & -- & -- & 2.79 & 5.07 \\
      & DexYCB \;\; +VOKE       & \textbf{1.75} & \textbf{2.60} & \textbf{1.29} & \textbf{59.1} & -- & -- & -- & -- & \textbf{1.27} & \textbf{4.40} \\
      & OakInk \;\; orig.       & 2.12 & 2.95 & 1.46 & 53.7 & -- & -- & -- & -- & 2.71 & 4.99 \\
      & OakInk \;\; +VOKE       & \textbf{2.03} & \textbf{2.83} & \textbf{1.40} & \textbf{54.9} & -- & -- & -- & -- & \textbf{1.60} & \textbf{4.31} \\
    \bottomrule
  \end{tabular}
  }
\end{table*}

\subsection{Main Results: Explicit Velocity and Acceleration Estimation}
\label{sec:main_kin}

We evaluate the accuracy of the motion vector predicted by PKM on both benchmarks and the dexterous in-hand manipulation simulation dataset. Our pose-derived baseline applies a second-order Savitzky--Golay fit with a five-frame window to each predicted pose sequence~\cite{savitzky1964smoothing}. Translations are fitted component-wise, while rotations are mapped to a local $SO(3)$ tangent space before fitting; analytic first and second derivatives yield $(v,\omega,a,\beta)$. We call this baseline \textbf{Pose Diff (SG)}.

\begin{figure*}[!t]
  \centering
  \includegraphics[width=0.8\textwidth]{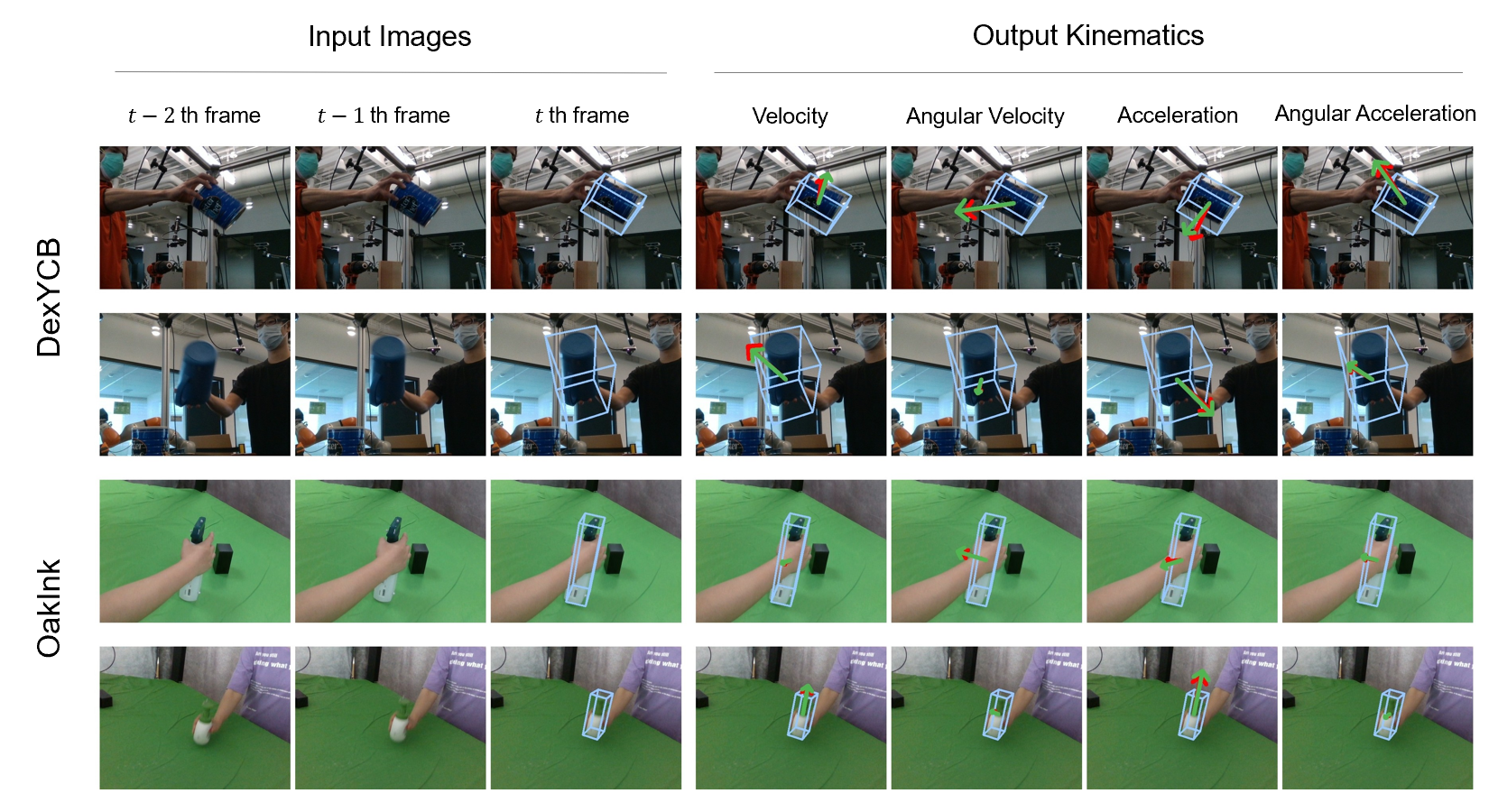}
  \caption{Visualization of VOKE on DexYCB and OakInk datasets. The three left columns show the last three input frames; the four right columns show estimates of linear and angular velocity and acceleration. Green arrows denote the reference motion quantities derived from ground-truth poses, red arrows denote VOKE estimates, and light-blue boxes visualize the 3D bounding boxes of the refined poses.}
  \label{fig:est_kin_open}
\end{figure*}

The quantitative results are shown in Table~\ref{tab:kin_est_all}. VOKE yields lower MEE and NRMSE than Pose Diff (SG) for all four motion quantities across all three datasets. Every VOKE NRMSE remains below 1, whereas Pose Diff (SG) exceeds 1 for every angular-velocity and acceleration quantity and for all four quantities in simulation. This consistent advantage supports estimating motion directly from temporal visual features rather than differentiating noisy pose predictions, answering \textbf{Q2} affirmatively. A qualitative visualization is shown in Fig.~\ref{fig:est_kin_open}.

\begin{table*}[!t]
  \caption{Object velocity and acceleration estimation on DexYCB, OakInk, and the dexterous in-hand manipulation simulation dataset. Here, $v$, $\omega$, $a$, and $\beta$ denote linear velocity, angular velocity, linear acceleration, and angular acceleration, respectively. Their MEE units are $m/s$, $rad/s$, $m/s^2$, and $rad/s^2$, respectively; NRMSE is dimensionless.}
  \label{tab:kin_est_all}
  \label{tab:kin_est_dexycb}
  \label{tab:kin_est_oakink}
  \label{tab:kin_est_sim}
  \centering
  \scriptsize
  \begin{tabular*}{\textwidth}{@{\extracolsep{\fill}}llcccccccc@{}}
    \toprule
    Dataset & Method & \multicolumn{4}{c}{MEE $\downarrow$} & \multicolumn{4}{c}{NRMSE $\downarrow$} \\
    \cmidrule(lr){3-6}\cmidrule(lr){7-10}
    & & $v$ & $\omega$ & $a$ & $\beta$ & $v$ & $\omega$ & $a$ & $\beta$ \\
    \midrule
    \multirow{2}{*}{DexYCB}
      & Pose Diff (SG) & 0.147 & 2.022 & 6.176 & 32.997 & 0.832 & 1.472 & 4.928 & 2.367 \\
      & VOKE & \textbf{0.087} & \textbf{0.965} & \textbf{1.307} & \textbf{19.832} & \textbf{0.259} & \textbf{0.468} & \textbf{0.871} & \textbf{0.984} \\
    \midrule
    \multirow{2}{*}{OakInk}
      & Pose Diff (SG) & 0.140 & 1.858 & 6.064 & 27.507 & 0.932 & 1.225 & 6.128 & 2.398 \\
      & VOKE & \textbf{0.065} & \textbf{0.553} & \textbf{0.845} & \textbf{12.225} & \textbf{0.317} & \textbf{0.456} & \textbf{0.778} & \textbf{0.898} \\
    \midrule
    \multirow{2}{*}{Simulation}
      & Pose Diff (SG) & 0.598 & 12.552 & 16.142 & 169.133 & 1.550 & 1.790 & 5.005 & 4.127 \\
      & VOKE & \textbf{0.107} & \textbf{4.277} & \textbf{7.112} & \textbf{27.498} & \textbf{0.353} & \textbf{0.516} & \textbf{0.823} & \textbf{0.992} \\
    \bottomrule
  \end{tabular*}
\end{table*}

\subsection{Ablation Studies}
\label{sec:ablation}

\subsubsection{Ablation on Pose Refinement Components}

To analyze how the proposed kinematics-guided refinement improves object pose estimation, we conduct an ablation study on the gSDF backbone to distinguish the contributions of generic temporal pose aggregation, the explicit motion vector, and the physics-inspired reference pose. We compare the following variants:

\begin{itemize}
    \item \textbf{VOKE w/o motion vector \& reference pose.} The pose refiner only takes the historical pose predictions $P_{t-k+1:t,\mathrm{net}}$ as input, without $V_t$ or the reference pose. This tests whether pose estimation improves merely by aggregating recent predictions.
    \item \textbf{VOKE w/o reference pose.} The pose refiner takes both the historical poses and the estimated motion vector $V_t$, but does not integrate $V_t$ into the reference pose $P_{t,\mathrm{ref}}$. This tests whether the explicit velocity and acceleration quantities alone are beneficial.
    \item \textbf{VOKE (full).} This complete model additionally receives the reference pose obtained by integrating the previous pose with $V_t$.
\end{itemize}

The results are summarized in Table~\ref{tab:ablation_pose_all}. Using pose history alone has only a marginal effect and is not uniformly beneficial on DexYCB. Adding the explicit motion vector improves all reported metrics on both datasets, and the full model performs best throughout. These consistent gains indicate that the motion-derived reference pose provides a useful second-order physical prior beyond direct access to $V_t$.

\begin{table}[t]
\caption{Ablation study of pose refinement components on the DexYCB and OakInk datasets.}
\label{tab:ablation_pose_all}
\label{tab:ablation_pose_dexycb}
\label{tab:ablation_pose_oakink}
  \centering
  \resizebox{\columnwidth}{!}{
  \begin{tabular}{lccccc}
    \toprule
    Method & \multicolumn{2}{c}{$CD_o\downarrow$} & $FS_o@5\uparrow$ & ADD-S$\downarrow$ & $E_o\downarrow$ \\
    \cmidrule(lr){2-3}
    & mean & median & & & \\
    \midrule
    \multicolumn{6}{c}{DexYCB} \\
    \midrule
    gSDF & 3.76 & 1.58 & 0.439 & 1.40 & 1.95 \\
    VOKE w/o motion \& ref.\ pose & 3.70 & 1.59 & 0.438 & 1.41 & 1.94 \\
    VOKE w/o ref.\ pose & 3.66 & 1.54 & 0.442 & 1.38 & 1.90 \\
    VOKE (full) & \textbf{3.50} & \textbf{1.43} & \textbf{0.456} & \textbf{1.35} & \textbf{1.80} \\
    \midrule
    \multicolumn{6}{c}{OakInk} \\
    \midrule
    gSDF & 6.51 & 1.52 & 0.414 & 1.55 & 2.24 \\
    VOKE w/o motion \& ref.\ pose & 6.45 & 1.52 & 0.415 & 1.55 & 2.22 \\
    VOKE w/o ref.\ pose & 6.21 & 1.49 & 0.418 & 1.53 & 2.18 \\
    VOKE (full) & \textbf{6.07} & \textbf{1.46} & \textbf{0.428} & \textbf{1.48} & \textbf{2.12} \\
    \bottomrule
  \end{tabular}
  }
\end{table}

\subsubsection{Ablation on Temporal Window Size $k$}

We study the effect of the number of input frames $k$ used by PKM and KPRM on both motion estimation accuracy and refined pose quality. Table~\ref{tab:ablation_k_kin_dexycb} reports the velocity and acceleration results for $k \in \{1, 2, 3, 5\}$ on DexYCB, and Table~\ref{tab:ablation_k_pose_dexycb} reports the corresponding refined pose quality. Increasing $k$ from 1 to 3 sharply reduces all motion-quantity errors and improves the refined pose. Extending the window to five frames yields only small additional gains for acceleration and $FS_o@5$; $CD_o$ and ADD-S remain unchanged, while velocity estimation and $E_o$ slightly degrade. We therefore use $k=3$ as the best accuracy--latency trade-off.

\begin{table}[t]
  \caption{Ablation on the temporal window size $k$: velocity and acceleration estimation accuracy on DexYCB. Here, $v$, $\omega$, $a$, and $\beta$ denote linear velocity, angular velocity, linear acceleration, and angular acceleration, respectively. Their MEE units are $m/s$, $rad/s$, $m/s^2$, and $rad/s^2$, respectively; NRMSE is dimensionless.}
  \label{tab:ablation_k_kin_dexycb}
  \centering
  \scriptsize
  \begin{tabular*}{\columnwidth}{@{\extracolsep{\fill}}cccccc@{}}
    \toprule
    Metric & $k$ & $v$ & $\omega$ & $a$ & $\beta$ \\
    \midrule
    \multirow{4}{*}{MEE $\downarrow$}
      & 1 & 0.671 & 3.720 & 13.693 & 90.610 \\
      & 2 & 0.150 & 1.774 & 2.817 & 62.089 \\
      & 3 & \textbf{0.087} & \textbf{0.965} & 1.307 & 19.832 \\
      & 5 & 0.090 & 0.971 & \textbf{1.210} & \textbf{18.718} \\
    \midrule
    \multirow{4}{*}{NRMSE $\downarrow$}
      & 1 & 1.717 & 1.923 & 7.198 & 5.205 \\
      & 2 & 0.516 & 0.958 & 3.950 & 2.993 \\
      & 3 & \textbf{0.259} & \textbf{0.468} & 0.871 & 0.984 \\
      & 5 & 0.263 & 0.470 & \textbf{0.859} & \textbf{0.979} \\
    \bottomrule
  \end{tabular*}
\end{table}

\begin{table}[t]
  \caption{Ablation on the temporal window size $k$: refined pose quality on DexYCB (gSDF + VOKE).}
  \label{tab:ablation_k_pose_dexycb}
  \centering
  \scriptsize
  \begin{tabular*}{\columnwidth}{@{\extracolsep{\fill}}cccccc@{}}
    \toprule
    $k$ & \multicolumn{2}{c}{$CD_o\downarrow$} & $FS_o@5\uparrow$ & ADD-S$\downarrow$ & $E_o\downarrow$ \\
    \cmidrule{2-3}
    & mean & median & & & \\
    \midrule
    1 & 3.72 & 1.58 & 0.440 & 1.40 & 1.95 \\
    2 & 3.55 & 1.50 & 0.447 & 1.37 & 1.89 \\
    3 & \textbf{3.50} & \textbf{1.43} & 0.456 & \textbf{1.35} & \textbf{1.80} \\
    5 & \textbf{3.50} & \textbf{1.43} & \textbf{0.458} & \textbf{1.35} & 1.81 \\
    \bottomrule
  \end{tabular*}
\end{table}

\begin{figure*}[!t]
  \centering
  \includegraphics[width=0.62\textwidth]{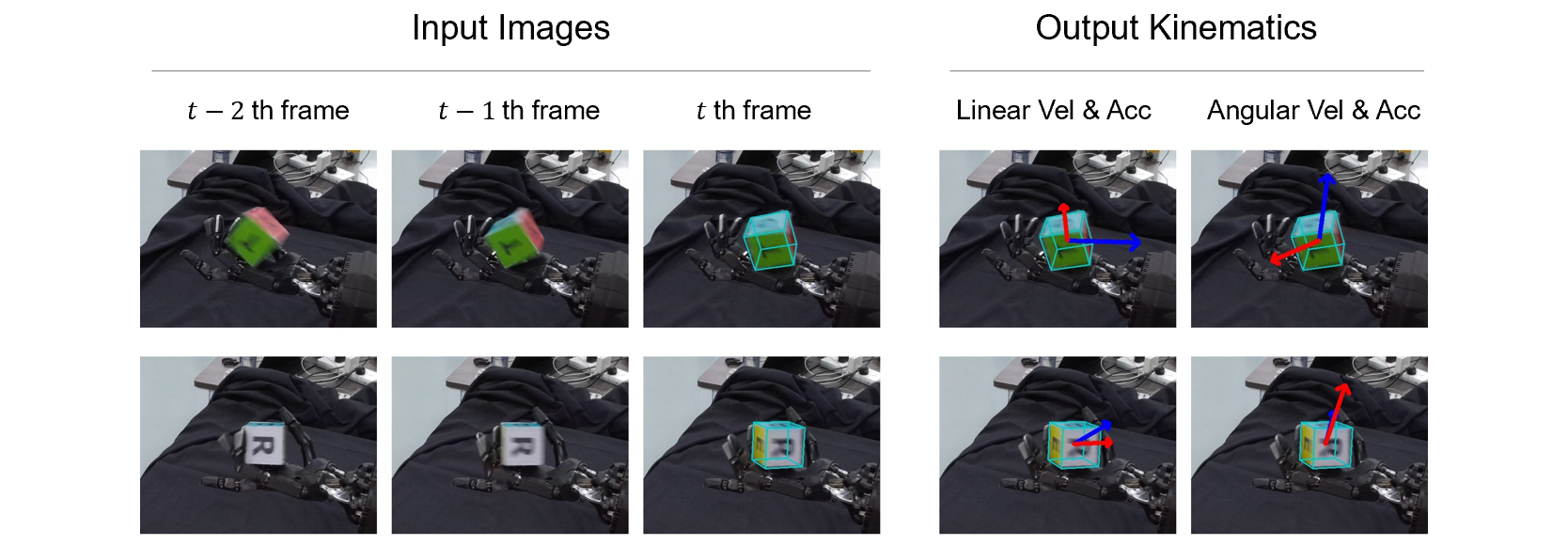}
  \caption{
  Online VOKE predictions during real-world deployment of the simulation-trained kinematics-augmented policy. The three left columns show consecutive input frames, and the two right columns visualize the predicted linear and angular velocity and acceleration. Blue arrows denote estimated velocity and red arrows denote estimated acceleration.
  }
  \label{fig:est_kin_real}
\end{figure*}

\subsection{Application Study: Dexterous In-Hand Manipulation with KARL}
\label{sec:application}

VOKE combines the refined pose with estimated velocity and acceleration to provide a complete object-centric kinematic state. We study this representation in dexterous in-hand manipulation, where continuous control, frequent contacts, occlusion, and rapidly changing motion make explicit dynamic observations particularly relevant.

\paragraph{Task and setup} We consider a dexterous in-hand cube reorientation task in which a cube is rotated to a target orientation within the workspace of a Shadow Dexterous Hand. The simulated task provides the quantitative comparison, after which we transfer the trained policy to the physical hand. During deployment, VOKE processes a live Azure Kinect RGB stream and supplies the predicted complete kinematic state to the policy before each action.

\paragraph{Simulated policy training and evaluation} We train both PPO variants in Isaac Sim with five independent seeds and identical proprioceptive and goal-related inputs. The history baseline receives $(P_{t-2},P_{t-1},P_t)$, whereas KARL receives $K_t=(P_t,V_t)$ with $V_t=(v_t,\omega_t,a_t,\beta_t)$; thus, motion is represented implicitly by pose history or explicitly by the current pose, velocity, and acceleration. For each checkpoint and object-state source, we run 200 paired held-out episodes per seed (1,000 per condition). All conditions share initial states, targets, randomization realizations, episode horizon, and success criterion, and use frozen policies with deterministic actions. We report the mean across the five training seeds.

\paragraph{Training performance} The kinematics-augmented policy converges faster, achieves a higher average episode reward (3699 vs.\ 3157), and attains more consecutive successes (13.29 vs.\ 11.19), indicating that the complete kinematic-state observation improves policy training.

\paragraph{Closed-loop evaluation with estimated observations} Training curves do not establish end-to-end performance when the object state is obtained from images. We therefore evaluate the frozen policies with paired simulator-GT and causal VOKE-predicted inputs (Table~\ref{tab:karl_eval}). The former isolates the observation design under oracle state, whereas the latter is our primary end-to-end test; their gap reflects sensitivity to perception error. All inputs share the coordinate frame, normalization, and time alignment used in training. In both settings, the kinematics-augmented policy achieves higher success and lower final rotation error than the pose-history baseline. The advantage is strongest with predicted observations, improving success by 10.7 percentage points while reducing rotation error by 3.85 degrees. Thus, despite perception noise, the complete state formed by pose, velocity, and acceleration remains beneficial in the closed loop.

\begin{table}[t]
  \centering
  \captionsetup{width=\columnwidth}
  \caption{Closed-loop dexterous in-hand cube reorientation evaluation with oracle and visually estimated object-state inputs. Success rate is reported in percent, and final rotation error in degrees. Each condition contains 1,000 paired held-out episodes (five training seeds and 200 episodes per seed). Results are averaged across the five training seeds.}
  \label{tab:karl_eval}
  \scriptsize
  \setlength{\tabcolsep}{2.6pt}
  \begin{tabular*}{\columnwidth}{@{\extracolsep{\fill}}llcc@{}}
    \toprule
    Source & Object-state input & \shortstack{Success \\ rate $\uparrow$} & \shortstack{Rot. \\ err. $\downarrow$} \\
    \midrule
    \multirow{2}{*}{Sim. GT}
      & History $P^{\mathrm{gt}}_{t-2:t}$ & 97.20 & 5.25 \\
      & Complete $K^{\mathrm{gt}}_t=(P^{\mathrm{gt}}_t,V^{\mathrm{gt}}_t)$ & \textbf{98.10} & \textbf{3.69} \\
    \midrule
    \multirow{2}{*}{VOKE pred.}
      & History $\hat{P}_{t-2:t}$ & 68.20 & 12.75\\
      & Complete $\hat{K}_t=(\hat{P}_t,\hat{V}_t)$ & \textbf{78.90} & \textbf{8.90} \\
    \bottomrule
  \end{tabular*}
\end{table}

\paragraph{Sim-to-real policy deployment} We deploy the frozen kinematics-augmented policy on the physical robot without additional policy fine-tuning. Privileged simulator object states are unavailable at deployment; instead, every incoming RGB frame is processed online by VOKE, and the resulting complete state $\hat{K}_t=(\hat{P}_t,\hat{V}_t)$ replaces the simulator state in the policy observation. Figure~\ref{fig:est_kin_real} visualizes predictions from this closed-loop deployment. The real-world result is qualitative evidence of end-to-end sim-to-real execution, while the paired simulation study provides the quantitative control comparison. Together, these experiments address \textbf{Q3} at complementary levels of evidence.


\section{Conclusion}

We presented VOKE, a kinematics-aware framework that equips existing vision-based object and hand-object pose estimators with a complete kinematic state. PKM predicts the motion vector of linear and angular velocity and acceleration from temporal visual features, while KPRM uses the resulting second-order motion prior to refine per-frame poses. Together, the refined pose and motion vector form the state consumed by KARL. VOKE improves pose estimation across three backbones and outperforms smoothed pose differentiation on DexYCB and OakInk. KARL demonstrates the value of this complete state for simulated dexterous in-hand manipulation and transfers the frozen policy to a physical robot without fine-tuning.

VOKE remains limited by its backbone: severe occlusion or extreme viewpoints can corrupt the estimated motion vector and reference-pose refinement. Acceleration, especially angular acceleration, is also difficult under rapid contacts and slipping. Future work should improve robustness and quantify how estimation errors affect real-world closed-loop manipulation.




\bibliographystyle{IEEEtran}
\bibliography{IEEEabrv,example}

\end{document}